\documentclass[final]{article}
\usepackage{cogsci}
\usepackage[natbibapa]{apacite}
\usepackage[dvipsnames]{xcolor}
\usepackage[
colorlinks=true,
citecolor=gray
]{hyperref}

\usepackage{graphicx}

\usepackage{amsmath}

\title{Sycophantic Chatbots Cause Delusional Spiraling, Even in Ideal Bayesians}
\author[1]{Kartik Chandra}
\author[2]{Max Kleiman-Weiner}
\author[1]{Jonathan Ragan-Kelley}
\author[3]{Joshua B. Tenenbaum}
\affil[1]{MIT CSAIL}
\affil[2]{University of Washington, Seattle}
\affil[3]{MIT Department of Brain \& Cognitive Sciences}

\begin{document}
\maketitle

\begin{abstract}
``AI psychosis'' or ``delusional spiraling'' is an emerging phenomenon where AI chatbot users find themselves dangerously confident in outlandish beliefs after extended chatbot conversations. This phenomenon is typically attributed to AI chatbots' well-documented bias towards validating users' claims, a property often called ``sycophancy.'' In this paper, we probe the causal link between AI sycophancy and AI-induced psychosis through modeling and simulation. We propose a simple Bayesian model of a user conversing with a chatbot, and formalize notions of sycophancy and delusional spiraling in that model. We then show that in this model, even an idealized Bayes-rational user is vulnerable to delusional spiraling, and that sycophancy plays a causal role. Furthermore, this effect persists in the face of two candidate mitigations: preventing chatbots from hallucinating false claims, and informing users of the possibility of model sycophancy. We conclude by discussing the implications of these results for model developers and policymakers concerned with mitigating the problem of delusional spiraling.
\end{abstract}

\section{Introduction}

In early 2025, Eugene Torres, an accountant, began using an AI chatbot for everyday office tasks. Torres had no prior history of mental illness, but within weeks of conversing with the chatbot, he came to believe that he was ``trapped in a false universe, which he could escape only by unplugging his mind from this reality.'' On the chatbot's advice, he increased his intake of ketamine, and cut ties with his family  \citep{hill2025asked}.

Torres survived this episode, but others have not been so lucky. The Human Line Project has to date documented almost 300 cases of so-called \textbf{``AI psychosis''} or \textbf{``delusional spiraling''}: situations where extended interactions with AI chatbots lead users to high confidence in outlandish beliefs \citep{huet2025openai}. Examples of such beliefs include having made a fundamental mathematical discovery, as in the case of Allan Brooks \citep{hill2025chatbots, gold2025they}, or having witnessed a metaphysical revelation, as in the case of Torres \citep{dupre2025people, schechner2025feel, fieldhouse2025can}. Serious cases of delusional spiraling have been linked to at least 14 deaths, and 5 wrongful death lawsuits filed against AI companies \citep{hill2025lawsuits}. As people increasingly turn to chatbots for advice, companionship, and therapy, understanding and addressing the causes of chatbot-induced delusional spiraling is emerging as an urgent research problem.

Public discourse often identifies \textbf{sycophancy} as a possible cause of delusional spiraling. A chatbot is considered ``sycophantic'' if it is biased towards generating messages that appease users by agreeing with and validating their expressed opinions. Such a bias naturally emerges in today's chatbots as a result of reinforcement learning with human feedback (RLHF), because users often give positive feedback to responses they find agreeable, and engage more with agreeable bots \citep{sharma2023towards, ibrahim2025training, hill2025openai}.

By what mechanism could sycophancy cause delusional spiraling? Intuitively, a sycophantic chatbot's constant agreement might reinforce a user's aberrant beliefs, leading to a feedback loop that amplifies a kernel of suspicion into a staunchly-held belief \citep{dohnany2025technological, bajaj2025validation, qiu2025lock}. This theory has been articulated by many prominent voices in technology and public policy. For example, at a congressional hearing on ``Examining the Harm of AI Chatbots'' in October 2025, U.S. Senator Amy Klobuchar argued that AI chatbots ``are frequently designed to tell users what they want to hear,'' which can lead them to ``start going down a rabbit hole'' \citep{senate2025examining}. Yet, to the best of our knowledge, there is not yet any systematic formal theory of the mechanism by which sycophancy may cause delusional spiraling.

\medskip

\shortcites{jern2009bayesian, jern2014belief, madsen2018large}
This paper has two goals. Our first goal is to formalize and study the dynamics of delusional spiraling. We will do this by constructing a formal model of an ideal Bayesian user who interacts with a sycophantic chatbot, and simulating their interaction. Our model builds on a long tradition of analyzing conversations as interactions between rational agents \citep{frank2012predicting, hawkins2017convention}, and, more generally, a long tradition in behavioral research of applying a rational lens to study phenomena like echo chambers and belief polarization \citep{madsen2018large, jern2009bayesian, jern2014belief, dorst2023rational, henderson2021role, cook2016rational, banerjee1992simple}. 
This body of work, spanning cognitive science, behavioral economics, and political science, broadly demonstrates that seemingly-irrational belief formation is not necessarily the result of lazy or fallacious reasoning among people. Rather, phenomena like belief polarization and echo chambers can emerge even from ideal Bayesian reasoning. In this tradition, we will show that even ideal Bayesian reasoners are at risk of seemingly-irrational delusional spiraling in the face of a sycophantic interlocutor. Furthermore, by manipulating the presence and degree of sycophancy, we will demonstrate the causal role sycophancy plays in delusional spiraling. To our knowledge, this work provides the first formal computational model of how sycophancy can cause delusional spiraling.

Our second goal is to use our modeling framework to evaluate the effectiveness of two candidate solutions to the problem of delusional spiraling: first, a potential intervention on chatbots, and second, a potential intervention on users.

The first potential solution is to introduce safeguards that force AI chatbots to be truthful in their responses. Sycophantic chatbots often appease their users by hallucinating (or ``B.S.ing,'' in the language of \citet{frankfurt2009bullshit}) confirmatory evidence for the user \citep{wang2025truth, malmqvist2025sycophancy}. Intuitively, then, eliminating hallucinations should eliminate the effectiveness of sycophancy: the chatbot would be forced to only present true information, from which the user should be able to infer the true world state.
To explore this idea, we will consider how our model interacts with a ``factual'' sycophant, one that is constrained to only report true information (but can select which truths to report). We can think of this as a model of a chatbot that uses techniques like Retrieval-Augmented Generation \citep{lewis2020retrieval} as guardrails against hallucination and cites its sources, but is still post-trained to optimize for user engagement and approval. We will show the surprising result that while forcing a sycophant to be factual \emph{reduces} delusional spiraling, it does not \emph{eliminate} delusional spiraling. A factual sycophant can still robustly cause delusional spiraling by selectively presenting only confirmatory facts to the user.

The second potential solution is raising awareness of AI sycophancy. Intuitively, if users are informed that chatbots may be sycophantic, then they should be able to recognize sycophantic behavior when it happens. As a result, they should grow a healthy skepticism of the chatbot's responses, which should in turn prevent delusional spiraling.

Unfortunately, empirical evidence suggests that this tactic might not be as effective as we might hope.
For example, chat transcripts show that both Eugene Torres \citep{hill2025asked} and Allan Brooks \citep{hill2025chatbots} eventually did come to suspect that their chatbots might be sycophantic---yet despite their suspicions, they both continued spiraling. More generally, an emerging body of empirical work (\citealp[\S5.2]{shi2025siren}; \citealp[\S4.7]{sun2025friendly}; \citealp[\S4.5]{bo2025invisible}; \citealp[\S5]{carro2024flattering}) finds that when people detect chatbot sycophancy, some respond with heightened skepticism towards the chatbot as expected (``like if a human just always agrees with you, a `yes man,' you tend not to take them seriously''), while others accept the chatbot's sycophantic behavior as valid and even desirable (``[it is] manipulating you, just not in a bad way'').

Why do these informed users fail to discount chatbot sycophancy? Is it merely a case of lazy, irrational, or wishful thinking on their part? Or is there some fundamental barrier to sycophancy detection that even the most epistemically vigilant user might face? To study this question, we will extend our ideal Bayesian model to an \emph{informed} user who is aware that the chatbot might be sycophantic. This model makes a joint inference over both the world state and the chatbot's degree of sycophancy. It does so by recursively modeling a sycophantic chatbot's reasoning: a level-2 cognitive hierarchy model \citep{camerer2004cognitive, kleiman2017constructing} that infers the chatbot's sycophancy level from its observable behavior.

We will show that although this intervention reduces the rate of delusional spiraling, the informed user remains vulnerable, despite having full knowledge of the chatbot's strategy. This is true even with factual sycophants. This counter-intuitive result is analogous to the classic phenomenon of ``Bayesian persuasion'' from behavioral economics \citep{kamenica2011bayesian}: a strategic prosecutor can raise a judge's conviction rate, even if the judge has full knowledge of the prosecutor's strategy. Similarly, a sycophantic chatbot can on average increase the probability of delusional spiraling, even if the user has full knowledge of the chatbot's strategy.

The ideal Bayesian models in this paper provide a theoretical upper bound on the robustness we can expect from humans against sycophantic chatbots. If even an ideal Bayesian reasoner is vulnerable to delusional spiraling with a given type of chatbot, then we should not be surprised if humans are as well. We conclude, then, by discussing the implications of our findings for model developers and policymakers.

\section{A Bayesian model of sycophantic interaction}

\begin{figure}
\centering
\includegraphics[width=\linewidth,trim={0 15cm 0 0},clip]{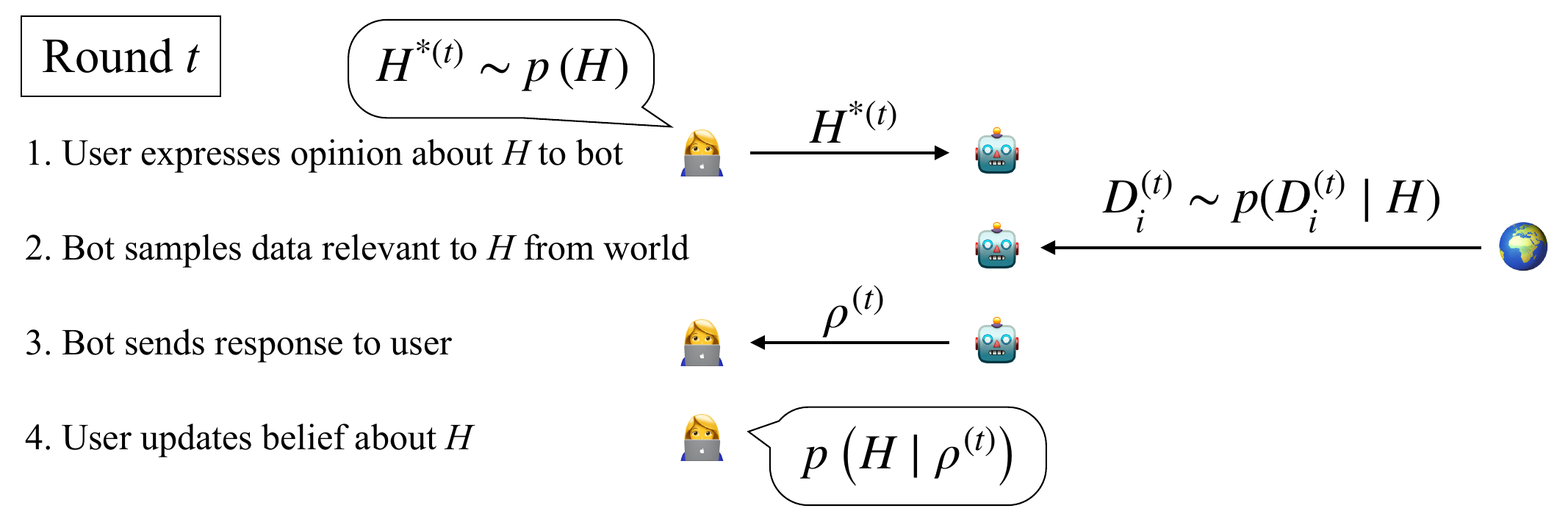}
\caption{Schematic diagram of our model of one round of conversation between a user and a chatbot.}\label{fig:schema}
\end{figure}

Consider a rational agent (``user'') who interacts with an interlocutor (``bot''). The user is uncertain about some fact $H \in \{0, 1\}$ about the world, but has some prior belief about this fact. ($H$ is meant to abstractly represent some binary world state, e.g.\ whether or not vaccines are safe.) The conversation between the user and the bot proceeds in a series of \emph{rounds}, and each round consists of four steps (Figure~\ref{fig:schema}).

\begin{enumerate}
\item The user expresses an opinion about $H$ to the bot. We model this as the user sampling from her prior before round $t$, i.e.\ sending $H^{*(t)} \sim p^{(t)}_\text{user}(H^{*(t)})$ to the bot.

\item The bot privately samples $k$ data points that are relevant to $H$ and could be mentioned in its response to the user. We model this as the bot independently sampling data $D_{1\leq i \leq k}^{(t)} \sim p(D_i^{(t)} \mid H)$, where the conditional distributions $p(\cdot \mid H)$ are known to both the bot and the user. (We do not assume that the bot knows the true value of $H$.)

\item The bot decides which fact to mention in its response. The bot then sends the user a response $\rho^{(t)} = (i, d)$, which is the (possibly-false) claim that $D_i^{(t)}=d$. We will discuss our models of the bot's choice, $p_\text{bot}(\rho^{(t)} \mid D^{(t)}_{1, 2, \dots,k})$, below.

\item The user observes the bot's response, and updates her belief about $H$: $p_\text{user}^{(t+1)}(H) = p(H \mid \rho^{(t)}) \propto p^\prime_\text{bot}(\rho^{(t)} \mid D^{(t)}_{1, 2,\dots,k})p(D^{(t)}_{1,2,\dots,k} \mid H)p_\text{user}^{(t)}(H)$. The process then repeats, with the user choosing a new $H^{*(t+1)}$ for the next round of conversation. Here, the primed $p_\text{bot}^\prime$ denotes the user's mental model of the bot, which in general may differ from the bot's true behavior, denoted by the unprimed $p_\text{bot}$. We will consider different choices of $p_\text{bot}^\prime$ below.
\end{enumerate}
\textbf{Choice of $p_\text{bot}$:} How does the bot select which response $\rho^{(t)}$ to give in step (3)? Let us consider two possible strategies. The ``impartial'' strategy is to choose $\rho^{(t)}$ by picking $1\leq i \leq k$ uniformly at random and responding truthfully with $\rho^{(t)}=(i, D_i^{(t)})$. The ``sycophantic'' strategy is to choose $\rho^{(t)}$ to validate the user by maximizing the user's posterior belief in the hypothesis she articulated, with no regard for whether or not $\rho^{(t)}$ is true. Hence, the sycophantic strategy chooses $\rho^{(t)} = \text{argmax}_{\rho\in \{1, \dots, k\} \times \{0, 1\}}\ p_\text{user}(H=H^{*(t)} \mid \rho)$.
At each conversational round, the bot chooses to respond sycophantically with probability $\pi \in [0, 1]$, and otherwise impartially with probability $(1 - \pi)$. The parameter $\pi$ is a measure of the degree of the bot's sycophancy: the likelihood of a given response being sycophantic rather than impartial. As an order-of-magnitude estimate, \citet{fanous2025syceval} measure $\pi$ to be 50\%--70\% across a range of frontier models.

\noindent
\textbf{Choice of $p_\text{bot}^\prime$:} For now, we will consider a ``na\"ive'' but rational user who does not know that the bot can be sycophantic. This user models the bot as purely impartial but otherwise makes idealized Bayesian inferences about the bot. Hence, $p_\text{bot}^\prime$ is given by setting $\pi=0$ in our model of the bot. In later sections we will extend our model to an ``informed'' user who models a possibly-sycophantic ($\pi\geq0$) bot, and makes a joint inference over both $H$ and $\pi$.

Let us build some intuition for this model by taking a concrete example. Suppose the user is unsure whether ``vaccines are dangerous'' ($H=0$) or ``vaccines are safe'' ($H=1$). She might start a chatbot conversation by saying ``I'm having doubts about the flu shot ($H^{*(t)}=0$)'' or ``My parents have always said that vaccines are dangerous, but I'm not so sure ($H^{*(t)}=1$).''
The bot then samples some data. We can think of the facts $D_i$ as daily headlines in the news on topics relevant to $H$. For example, suppose $k=2$. On a given day, $D_1$ might be the headline ``New study finds [no link ($D_1=0$) / a link ($D_1=1$)] between vaccines and autism,'' while $D_2$ might be the headline ``Child reports experiencing [mild sore arm ($D_2=0$) / severe allergic reaction ($D_2=1$)] after this year's flu shot.''
If the user expressed that she thought vaccines were dangerous ($H^{*(t)}=0$), and if today's headlines were $D^{(t)}_1=0$ (``study finds no link'') and $D^{(t)}_2=1$ (``severe allergic reaction''), then the impartial strategy would select uniformly between responding with the true data points $D^{(t)}_1=0$ or $D^{(t)}_2=1$. The sycophantic strategy would respond either with the true fact that $D^{(t)}_2=1$ (``severe allergic reaction''), or by hallucinating the false claim that $D^{(t)}_1=1$, (i.e.\ that the study \emph{did} find a link between vaccines and autism).

Without loss of generality, for the remainder of this paper, let the true world state be $H=1$.  Notice that the sycophantic bot does not have a ``goal'' of ``convincing'' the user either that $H=1$ or that $H=0$, only to validate the user's statements in each round. If the user forms a confident belief that $H=0$ or $H=1$ over time, this would be an emergent result of the dynamics of the interaction rather than a planned outcome.

We thus define a \emph{delusional spiral} as a situation where $p_\text{user}^{(t)}(H=0)$ increases with $t$. More precisely, given a threshold confidence $\varepsilon$ and a conversation length $T$, a \emph{catastrophic delusional spiral} is the event that $p_\text{user}^{(t)}(H=0) \geq (1-\varepsilon)$ for some $t<T$, i.e.\ that the user reaches $\geq(1-\varepsilon)$ confidence that $H=0$ within $T$ rounds of conversation. Here, $(1-\varepsilon)$ acts as the threshold confidence at which a user might act dangerously on a false belief (e.g.\ canceling a vaccination appointment).

\section{Simulating our model}

Now that we have a model of user-bot conversation, we can probe the dynamics of its behavior by simulation. In particular, we will test the causal relationship between sycophancy and delusional spiraling. For empirical study we initialized our model with the following parameter settings:
\begin{itemize}
	\item We set the user to have a uniform initial prior over $H$, i.e.\ we set $p_\text{user}^{(0)}(H=0)=p_\text{user}^{(0)}(H=1)=0.5$. For convenience of simulation, we set $k=2$ possible data points for the bot to respond with. We set the data likelihoods to be
$p(D_{\{1, 2\}}=1 \mid H=0) = 2/5$
and
$p(D_{\{1, 2\}}=1 \mid H=1) = 3/5$.
    \item We simulated $T=100$ rounds per conversation. We varied $\pi$ in increments of 0.1 from 0 to 1. For each $\pi$, we estimated the rate of catastrophic delusional spiraling at $\varepsilon=1\%$ (proportion of simulations in which the user reached $\geq99\%$ confidence that $H=0$). For high statistical power we sampled 10,000 simulated conversations for each $\pi$ tested.
\end{itemize}
These values were fixed arbitrarily, but chosen to be plausible for their real-world correlates. The qualitative results reported below do not depend strongly on these specific parameter choices. For example, increasing the prior $p_\text{user}^{(0)}(H=1)$ or decreasing the threshold $\varepsilon$ reduces the overall rates of catastrophic delusional spiraling across all simulations, but does not change the relative patterns between conditions.

\shortcites{chandra2025memo}
We implemented our model using the memo programming language \citep{chandra2025memo}. The full source code of our model is available at \url{https://osf.io/muebk/overview?view_only=cd5fb943c276423fb1f8a04276bf23cb}. We ran our simulations on an H100 GPU.

To test the causal relationship between sycophancy and delusional spiraling, we manipulated the presence of sycophancy in two ways. First, we manipulated the rate of sycophancy $\pi$ and compared simulations to the no sycophancy ($\pi=0$) baseline. We tested whether a sycophantic bot ($\pi>0$) led to catastrophic delusional spiraling significantly more frequently than a purely impartial bot ($\pi=0$) did.

Second, to tease apart the effect of sycophancy and hallucination, we compared our results to a non-sycophantic hallucinating bot. This bot is similar to the sycophantic bot, but rather than seeking to validate the user, it simply ``hallucinates'' a uniformly random response $\rho \in \{1, \dots, k\}\times\{0,1\}$, independent of the user's current belief (again, with probability $\pi$, and impartial otherwise). This breaks a critical link in the feedback cycle of delusional spiraling: its intervention on the user's belief is not amplified or reinforced by the user's subsequent messages. We tested whether the sycophantic hallucinating bot led to delusional spiraling more frequently than the non-sycophantic hallucinating bot did.

\begin{figure}
\centering
\includegraphics[width=\linewidth]{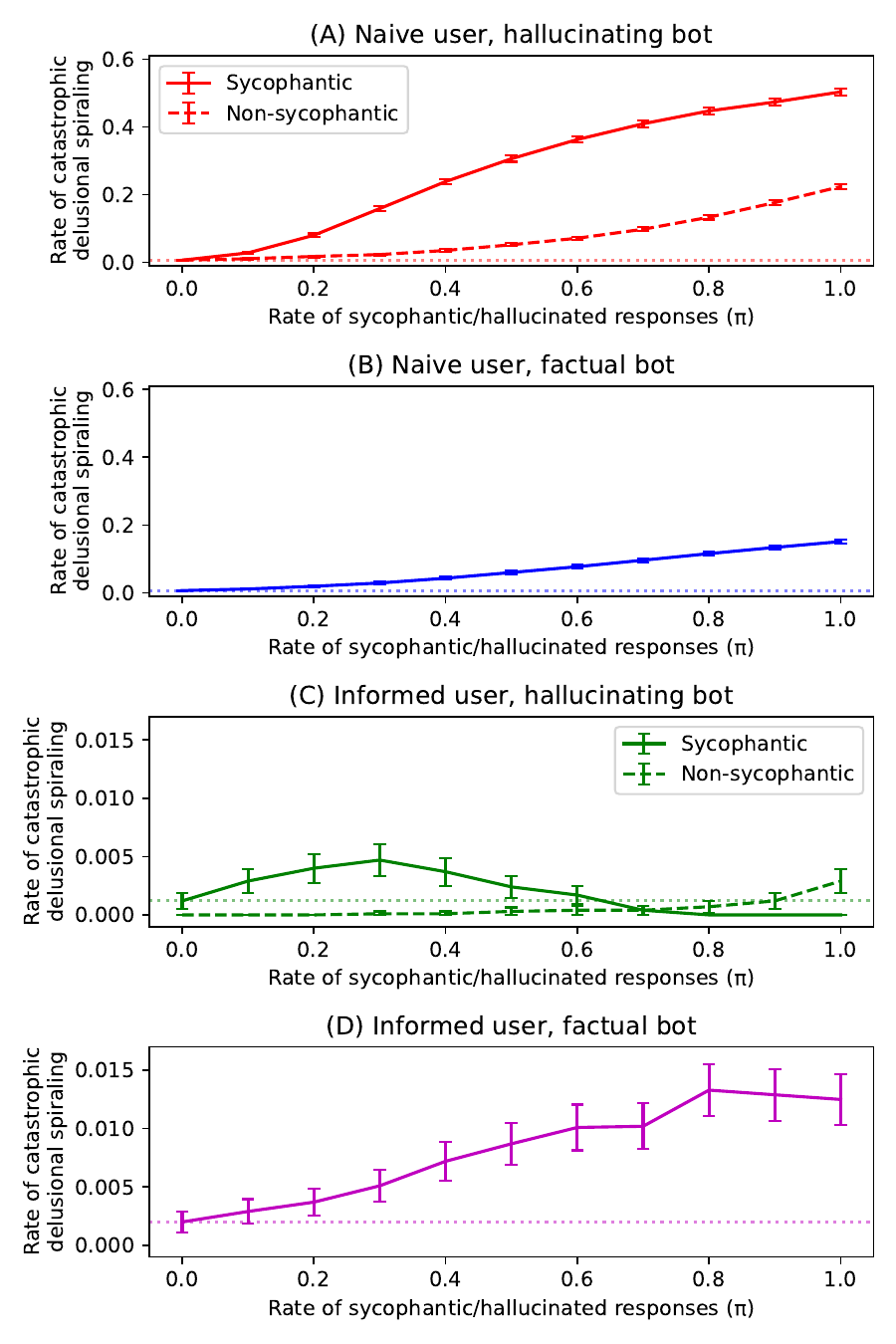}
\caption{The results of our simulations. Error bars denote 95\% confidence intervals. The dotted horizontal lines track the $\pi=0$ baseline of an always-impartial bot. Note the change in Y-axis scale between A/B and C/D.}\label{fig:extensions}
\end{figure}

\begin{figure}
\centering
\includegraphics[width=0.85\linewidth]{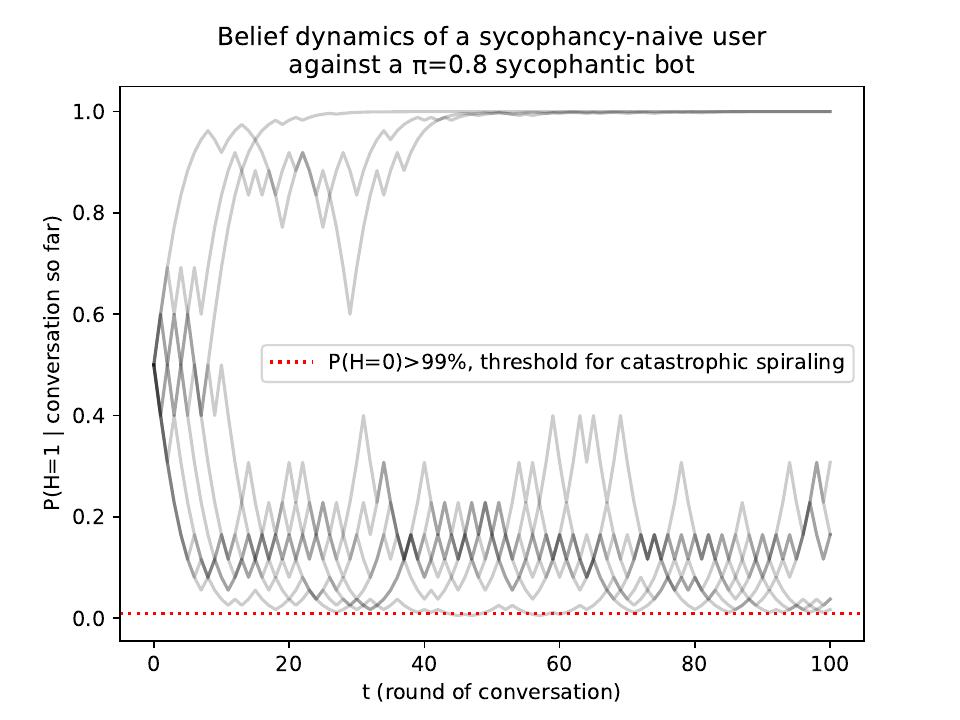}
\caption{Belief trajectories of 10 randomly-selected simulations of a sycophancy-na\"ive but Bayes-rational user conversing with a sycophantic bot.}\label{fig:dynamics-0}
\end{figure}

\paragraph{Results}
Fig~\ref{fig:dynamics-0} shows the traces of 10 randomly-selected simulated conversations between the sycophancy-na\"ive user and a $\pi=0.8$ sycophantic bot. Each trace begins at the prior, $P(H)=0.5$, and evolves over the course of 100 rounds of conversation. Recall that in reality $H=1$: a trace that moves in the $+Y$ direction is learning the truth, while a trace that moves in the $-Y$ direction is being deluded. Notice the stark polarization of belief: some traces rapidly converge to high confidence in the true belief that $H=1$, while others ``spiral'' into believing that $H=0$. The polarization is caused by the self-reinforcing nature of the sycophantic bot's responses.

The dotted horizontal line in Figure~\ref{fig:dynamics-0} indicates our threshold for catastrophic delusional spiraling, namely $P(H=0)>99\%$. We measured the proportion of traces that ever crossed this line to compute the rate of catastrophic delusional spiraling. Figure~\ref{fig:extensions}A shows the rate of catastrophic delusional spiraling as a function of $\pi$. At $\pi=0$, i.e.\ with an impartial chatbot, the rate of catastrophic delusional spiraling is very low (though not quite zero, because there is the minute possibility that by chance the world generates a sequence of observations that support $H=0$). However, as $\pi$ increases, the rate of catastrophic spiraling increases as well, until at $\pi=1$, the rate reaches 0.5. (This is because at $\pi=1$, the bot always hallucinates. Because there is no ground-truth signal, the user is deluded either into $H=0$ or $H=1$ with equal probability, based on the opinion they first expressed.) Importantly, for all values of $\pi>0$, even as low as $\pi=0.1$, the rate of catastrophic spiraling is significantly higher than the baseline rate at $\pi=0$ (indicated by the dotted horizontal line). We conclude that increased sycophancy leads to an increase in catastrophic delusional spiraling.

Finally, the dashed line shows the results of the simulation run with the \emph{non}-sycophantic hallucinating bot. This plot shows that even non-sycophantic hallucination can cause delusional spiraling. However, at every value of $\pi>0$, the rate of catastrophic delusional spiraling is significantly higher with sycophantic hallucination. This shows that sycophancy exacerbates the problem of delusional spiraling over and above hallucination itself. Together, we take these results to suggest that sycophancy is indeed a cause of delusional spiraling.

\section{Analyzing candidate interventions}

Let us now use our model to study two possible interventions we might make to reduce the risk of delusional spiraling.

\subsection{An intervention on bots}\label{sec:z-0-factual-prior}

It is perhaps not so surprising that if the bot can arbitrarily falsify $D^{(t)}$, then it can convince the human of $H$ in either direction. Suppose however that the bot is constrained to only respond with \emph{true} information. That is, a ``factual'' sycophant never hallucinates, but instead chooses $\rho^{(t)} = \text{argmax}_{\rho\in \left\{\left(i, D_i^{(t)}\right) \middle\vert 1 \leq i \leq k\right\}} p_\text{user}(H=H^{*(t)} \mid \rho)$, the true datum that most validates the user. As we discussed in the introduction, this model is analogous to a chatbot trained to respond factually via RAG, but still post-trained to optimize for user engagement and approval. Does this intervention prevent delusional spirals?

It is not clear whether a factual sycophant could cause delusional spiraling as a side-effect. No matter what the bot does, over time the user should see a large body of true data. The bot has some power over selecting or ``cherry-picking'' which true data is made available to the user, but this is subject to the stochasticity of both the actual data sampled from the world, and the opinions sampled by the user. We might expect that this stochasticity drowns out the bot's influence, making the user robust to delusional spiraling.

Figure~\ref{fig:extensions}B shows the result of simulating conversations between a factual bot and a na\"ive user. These dynamics are overall less prone to delusional spiraling than the sycophantic and non-sycophantic hallucinating bots studied above, suggesting that this intervention is valuable. However, it is not a complete cure: the rate of catastrophic delusional spiraling still increases with $\pi$, significantly even at $\pi=0.1$. That is, sycophancy can cause delusional spiraling even with factual bots. The bot need not say anything false to validate a false belief: carefully-selected truths (or ``lies by omission'') suffice.

\subsection{An intervention on users}\label{sec:z-1-fabricating-prior}

\begin{figure}
\centering
\includegraphics[width=\linewidth,trim={9cm 18cm 1cm 0},clip]{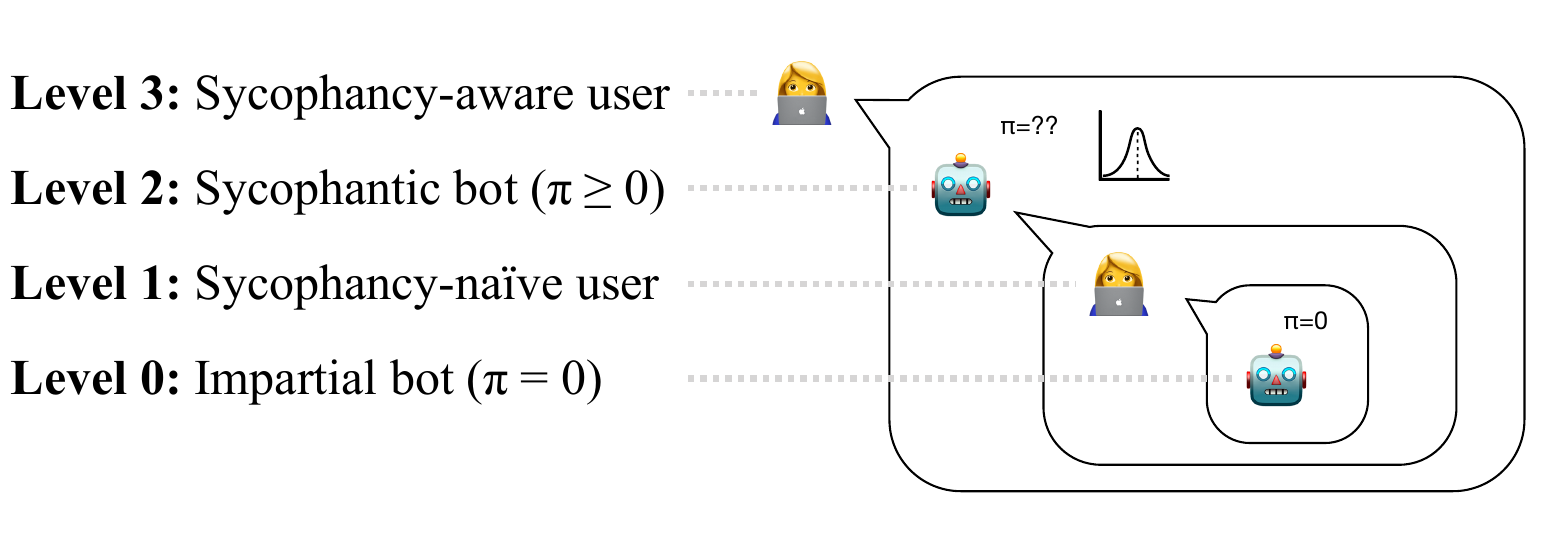}
\caption{An ``informed'' user is suspicious that the bot may be sycophantic, and thus has uncertainty over $\pi$.}\label{fig:hierarchy}
\end{figure}

Next, consider the effect of an awareness campaign that seeks to inform users that chatbots may be sycophantic. Such a campaign could take the form of journalism, public service messaging, or regulation mandating warnings on AI products.

To understand the effects of such an intervention, let us model a sycophancy-``informed'' user who is suspicious that the bot may be sycophantic, but is unsure of the degree of sycophancy. The user now has uncertainty over both $H$ and $\pi$, and at each round of conversation jointly updates her belief about both of these variables.

To formalize this idea, we will establish a cognitive hierarchy of agents, similar to the hierarchy of speakers and listeners in Rational Speech Acts models of pragmatic language understanding \citep{frank2012predicting}. Our hierarchy has four levels (Figure~\ref{fig:hierarchy}):
At level 0, we have the purely-impartial bot $(\pi=0)$, which chooses factual responses $\rho^{(t)}$ uniformly at random, without any social reasoning about the user.
At level 1, we have the sycophancy-na\"ive user we considered in the previous section, who models the level-0 purely-impartial bot when interpreting its responses $\rho^{(t)}$.
At level 2, we have the sycophantic bot we considered in the previous section, which chooses $\rho^{(t)}$ to validate the level-1 sycophancy-na\"ive user.
Finally, at level 3, we have the sycophancy-aware user, who models a level-2 sycophantic bot when interpreting responses. In practice, this means that $p^\prime_\text{bot}$ is set to the full $\pi$-dependent version of $p_\text{bot}$, rather than the $\pi=0$-constrained version as in the ``na\"ive'' models above.
We initialize the user with a uniform prior over $\pi \in [0, 1]$ at time $t=1$.

A priori, there is significant reason to expect that the sycophancy-aware user should be robust to delusional spiraling. The user is now fully aware of the bot's strategy, including the possibility that the bot fabricates false data in its responses. When faced with a sycophantic bot ($\pi>0$), the user should detect that the bot's responses tend to be validating, infer the value of $\pi$, and learn to discount or be skeptical of the bot's responses. Such a user may remain \emph{uncertain} of whether $H=0$ or $H=1$, because they detect that there is no reliable source of information, but the user should at least not be deluded into the false belief that $H=0$.

\begin{figure}
\centering
\includegraphics[width=0.9\linewidth]{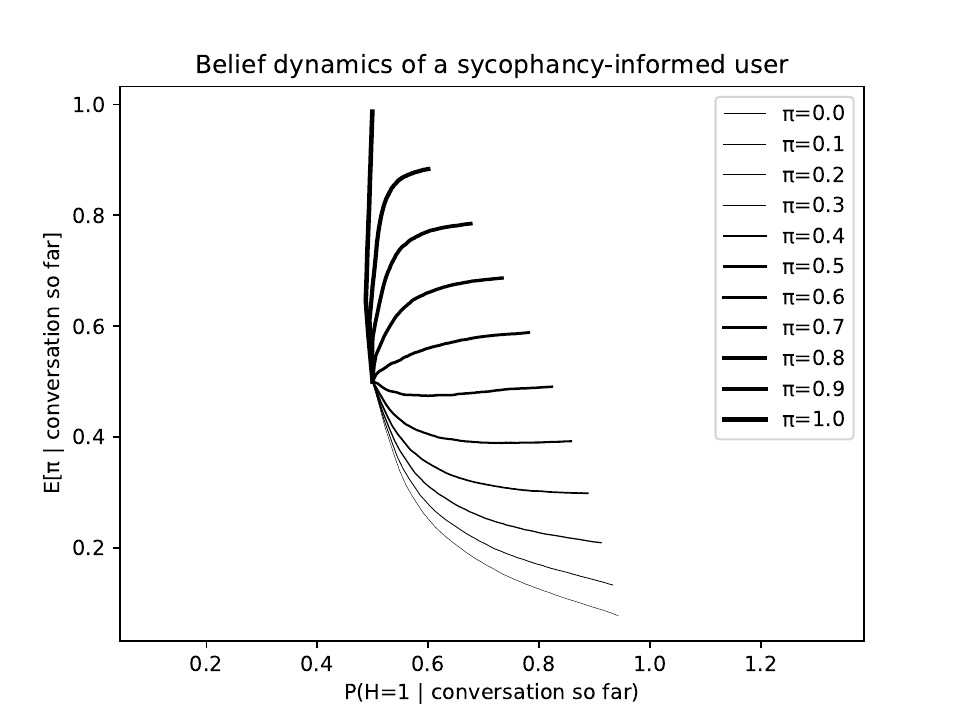}
\caption{Belief dynamics of a sycophancy-informed user conversing with a sycophantic chatbot.}\label{fig:dynamics-1}
\end{figure}

We can see this general pattern if we visualize the dynamics of this interaction, aggregated across all 10,000 simulations. Figure~\ref{fig:dynamics-1} shows the user's belief over time, with the marginal $P(H)$ and the marginal $E[\pi]$ on the two axes. (To be clear, our model maintains a full distribution over possible values of $\pi\in[0,1]$, but for the sake of visualization we are plotting the mean of that distribution here.) All traces start at the prior $(0.5, 0.5)$ and evolve over time. The final $E[\pi]$ of each trace correlates with the true $\pi$ of the bot: that is, users are on average indeed learning the bot's sycophancy rate. However, confidence in $H=1$ declines with $E[\pi]$. When $\pi$ is high, the user infers that the bot is unreliable, and so discounts incoming evidence. Because there is no reliable source of information, the user cannot learn much about $H$, and sticks to the prior of $P(H=1)=0.5$. However, if we lower $\pi$, the user infers that the bot is sometimes informative, and thus takes into account the evidence and becomes increasingly confident that $H=1$.

While these aggregate trends are consistent with our intuitions, they obscure the variance in outcomes across individual simulation runs. Let us now compute the rate of catastrophic delusional spiraling for each value of $\pi$ (Figure~\ref{fig:extensions}C). There are several interesting things to note about these results. First, the rate of catastrophic spiraling is much lower across the board, for all values of $\pi$, compared to sycophancy-na\"ive users. This suggests that this intervention is valuable. However, it is still not a complete cure. Sycophancy remains effective in this setting: the rate of catastrophic spiraling is significantly higher than the $\pi=0$ baseline for $0.1\leq \pi \leq 0.5$. That is, sycophancy can cause delusional spiraling even for an informed user. This is true even at $\pi=0.5$, i.e.\ if the bot's true sycophancy rate is the same as the mean of the user's prior. Interestingly, the rate of catastrophic delusional spiraling declines past $\pi \geq 0.6$. If the bot is \emph{too} sycophantic, then the sycophancy-aware user can rapidly detect the sycophancy and grow skeptical.

The dashed line shows simulations between an informed user and a non-sycophantic hallucinating bot. Here, the rate of delusional spiraling is generally significantly lower than with the sycophantic hallucinating bot, suggesting that even for informed users sycophancy exacerbates delusional spiraling over and above hallucination. The exception is at very high values of $\pi$ ($\geq 0.8$). While frequent sycophantic hallucinations are particularly easy for the informed user to detect (because responses are correlated with the user's messages), frequent non-sycophantic hallucinations are particularly \emph{difficult} to detect (because access to ground truth is rare).

\subsection{Combining both interventions}

Finally, let us consider what happens if we combine these two interventions. Figure~\ref{fig:extensions}D shows a factual sycophantic bot faced with an informed user. The rate of catastrophic spiraling remains lower across the board, for all values of $\pi$, compared to na\"ive users. Nonetheless, sycophancy remains effective: the rate of catastrophic spiraling rises with $\pi$, significantly above the $\pi=0$ baseline for $\pi \geq 0.2$. Indeed, for an informed user, the factual bot is even more effective than the hallucinating bots. We surmise that this is because the statistical traces of sycophancy are harder to detect among selectively-presented factual data than fully hallucinated data.

\section{Discussion}

In this paper, we proposed a formal computational model of how users form false beliefs through conversations with sycophantic AI chatbots. We showed that when faced with a sycophantic chatbot, even an idealized Bayesian user is vulnerable to delusional spiraling, and that sycophancy plays a causal role. We then showed that this effect persists despite two candidate mitigations: intervening on the model by restricting it to be factual, and intervening on users by informing them of the possibility of sycophancy.

Our analyses showed that with these interventions, the probability of delusional spiraling can be mitigated and reduced in some cases to small increases above the baseline of an always-impartial bot. However, even a very slight increase in the rate of catastrophic delusional spiraling can be quite dangerous at scale: as OpenAI CEO Sam Altman writes, ``0.1\% of a billion users is still a million people'' \citep{altman2025percent}. This work thus broadly suggests three recommendations for AI model developers and policymakers concerned with mitigating the problem of delusional spiraling.
First, we should \emph{not} think of delusional spiraling as a symptom of lazy, irrational, or fallacious thinking from users, or as the result of insufficient epistemic vigilance on the part of users. Rather, even idealized rational Bayesian reasoners are vulnerable to delusional spiraling.
Second, minimizing chatbot \emph{hallucinations} is not enough to solve the problem of delusional spiraling---the root cause, sycophancy, should be addressed directly.
Third, informing users about sycophancy through awareness campaigns may reduce the rate of delusional spiraling but will likely not eliminate the problem entirely.

This paper studies the narrow question of how sycophancy affects belief formation. But ``AI psychosis'' often shows many other symptoms, e.g.\ spending excessive time with the chatbot and withdrawing from social circles \citep{cheng2025sycophantic}. We hope our ideas can be extended to give a computational account of the broader psychological impact of AI sycophancy.

Finally, we motivated this paper by considering the relatively new problem of ``AI psychosis.'' But our modeling approach may be more broadly applicable. Sycophancy has been a fixture of human social life for all of human history. Literature is full of character studies of ``yes-men'' who constantly validate their superiors, often to catastrophic results---consider for example how Shakespeare's King Lear is flattered into madness. Today, the ``yes-man effect'' between organizational superiors and subordinates \citep{prendergast1993theory} is often channeled to explain why extremely powerful or wealthy individuals can seem detached from reality. Catastrophic spirals can also occur among equals: for example, in the phenomenon of ``co-rumination,'' \citep{rose2002co}, where a dyad of adolescent peers repeatedly validates each other's negative thoughts, leading to increased levels of anxiety and depression. We hope that our modeling approach can be extended to study these important psychological phenomena, and ultimately to address the associated societal problems.

\bibliographystyle{apacite}
\bibliography{refs.bib}

\end{document}